\documentclass[12pt]{article}
\usepackage[margin=1in]{geometry}
\usepackage{graphicx}
\usepackage{hyperref}
\usepackage{amsmath}
\usepackage{caption} 
\usepackage{etoolbox}
\apptocmd{\thebibliography}{\raggedright}{}{}
\usepackage[capitalise]{cleveref}
\usepackage{setspace}
\title{siForest: Detecting Network Anomalies with Set-Structured Isolation Forest}
\author{Christie Djidjev \\
	 {Department of Computer Science} \\
	{University of Texas at Austin}
}
\date{}
\begin{document}
	
	\maketitle
	
	\begin{abstract}
		As cyber threats continue to evolve in sophistication and scale, the ability to detect anomalous network behavior has become critical for maintaining robust cybersecurity defenses. Modern cybersecurity systems face the overwhelming challenge of analyzing billions of daily network interactions to identify potential threats, making efficient and accurate anomaly detection algorithms crucial for network defense. This paper investigates the use of variations of the Isolation Forest (iForest) machine learning algorithm for detecting anomalies in internet scan data. In particular, it presents the Set-Partitioned Isolation Forest (siForest), a novel extension of the iForest method designed to detect anomalies in set-structured data. By treating instances such as sets of multiple network scans with the same IP address as cohesive units, siForest effectively addresses some challenges of analyzing complex, multidimensional datasets. Extensive experiments on synthetic datasets simulating diverse anomaly scenarios in network traffic demonstrate that siForest has the potential to outperform traditional approaches on some types of internet scan data.
	\end{abstract}
	
	\section{Introduction}
	
	Protecting networks against cyber-attacks requires the correct identification of any potential weaknesses that could be exploited by attackers. Such vulnerabilities, collectively known as \emph{attack surface},  may include open ports, exposed services, misconfigured systems, and third-party integrations. The growing use of interconnected systems and digital technologies, using AI-driven attacks, and applying zero-day exploits leads to a larger attack surface and makes its identification harder. Hence, the use of automated systems capable of continuously monitoring, assessing, and mitigating vulnerabilities in real-time becomes essential. 
	
	One way to identify attack surface vulnerabilities is by using machine learning (ML) techniques. ML can detect \emph{anomalies} in the network traffic data, which may point to potential issues that need to be addressed. However, finding such anomalies is challenging due to the high-dimensional and variable nature of network data. Network scans often include multiple features per IP address, with significant variation in both quantity and type. 
	
	Machine learning techniques, such as clustering-based methods (e.g., k-means \cite{Jin2010}, DBSCAN \cite{ester1996density}) and tree-based algorithms (e.g., Isolation Forest \cite{liu2008isolation}, Random Forest \cite{ho1995random} anomaly detection), are widely used to identify patterns that deviate from expected behavior. 
	
\begin{figure}
	\centering
	\includegraphics[width=0.7\textwidth]{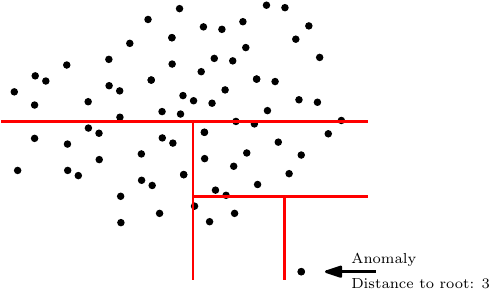}
	\caption{Illustration of how the distance to the root in an Isolation Forest tree is used to detect outliers. Shorter distances correspond to points that are easier to isolate and are therefore more likely to be anomalous.}
	\label{fig:iforestex}
\end{figure}

	\textit{Isolation Forest (iForest)}, which is one of the most scalable algorithms in this class, isolates anomalies by recursively splitting the data, and using the number of splits required to isolate a point as a measure of its anomaly score (Figure~\ref{fig:iforestex}). Specifically, it constructs an ensemble of binary trees (a \textit{forest}) by randomly selecting a feature and a split value within the feature's range at each node. The process continues recursively until  a single point is left (the point is isolated) or a maximum tree depth is reached. Points closer to the root of a tree (requiring fewer splits for isolation) are considered more anomalous, while those deeper in the tree (requiring more splits) are likely part of dense clusters. 
	
	However, this method shows limitations when applied to more complex types of data such as network scan data. iForest assumes fixed-length feature vectors and numerical inputs, which makes it less effective for set-structured data where features like ports and services vary across IPs. 
	
	To address these challenges, this paper introduces siForest, a variation of iForest designed specifically for set-structured data. The proposed method preserves the structure of the original data, and specifically the contextual relationships between ports and services under each IP. We experimentally compare siForest with two other implementations based on the original iForest implementation. The paper concluded with a discussion of the results and a list of tasks for further improvements.

	\section{Related Work}
	
	Network security analysts increasingly rely on anomaly detection to identify threats in complex environments. The theoretical foundations for modern detection methods \cite{chandola2009anomaly}, particularly those focusing on pattern recognition in security contexts, have evolved to address challenges in dynamic, large-scale networks where traditional approaches often fail \cite{ahmed2016survey}. Statistical methodologies \cite{markou2003novelty} have proven especially effective in environments processing terabytes of network traffic daily while maintaining high accuracy rates.
	
	The integration of machine learning has fundamentally changed anomaly detection capabilities, especially in analyzing high-dimensional data. While early research established core principles for identifying rare events \cite{domingos2012few}, recent advances in deep learning have revolutionized network traffic analysis through multi-layered pattern recognition \cite{yu2018networks}. Graph-based approaches \cite{akoglu2015graph} are particularly good at modeling complex  relationships between objects of interest, using advanced node feature learning techniques \cite{zhao2019learning}.
	
	Within cybersecurity applications, these methods demonstrate better effectiveness against zero-day exploits and massive-scale traffic analysis. Research into generalization capabilities across different network behaviors \cite{sommer2010outside} has revealed important insights for practical deployment. Deep hierarchical representations \cite{bengio2013representation} prove especially effective at capturing subtle attack patterns, leading to broader applications across multiple security domains \cite{lecun2015deep}.
	
	Isolation Forest \cite{liu2008isolation} is a more recent unsupervised machine learning algorithm particularly good  at processing high-dimensional data. It's main advantages are the high computational efficiency, while maintaining good accuracy. Further work \cite{hariri2019extended} improved the algorithm's resilience against complex attack patterns, and practical implementations \cite{liu2012isolation} demonstrated its effectiveness in production environments. Recent benchmarking studies \cite{zhu2022synthetic} analyzed the performance improvements across diverse scenarios.
	
	Recent advances in graph-based detection have yielded promising results in network security applications. While earlier work explored basic graph structures for relationship modeling \cite{akoglu2015graph}, the development of graph convolutional networks \cite{kipf2016semi} has dramatically improved detection accuracy in real-world security scenarios. These improvements, combined with advanced clustering techniques \cite{xie2017unsupervised}, demonstrate particular effectiveness in identifying coordinated attack patterns across distributed networks.
	
	Since real data can be hard and expensive to obtain, synthetic data generation has become instrumental in validating modern detection approaches. Building upon foundational work in dataset generation \cite{chandola2011data}, researchers have developed increasingly sophisticated evaluation frameworks. The introduction of Deep Sets \cite{zaheer2017deep} provided crucial tools for analyzing more complex security events, while new techniques  \cite{aggarwal2017outlier} address the challenges of high-dimensional, sparse data environments.
	
	\section{Methods}
	This section describes the data processing and machine learning methods  used to detect anomalies in network scan data. We first discuss the structure of the data as it determines the type of data preprocessing required. Then we describe two feature engineering methods that we use as well as the new siForest algorithm and its implementation. 
	
	\subsection{Censys Internet Scan Data}
	Censys is a company and internet intelligence platform designed to index and analyze data gathered from continuous, large-scale scans of the internet. For the entire public IPv4 address space, it scans for all available IP and port combinations with automated protocol identification. This process produces a precise snapshots of the current state of the internet. A typical dataset consists of a set of scans and each scan includes an IPv4 address, which is the unique identifier of the scanned device, a list of 
	services--applications or protocols available on the device, and a list of the corresponding network ports. 
	
	This dataset is categorical, consisting primarily of lists. Moreover,  objects in the lists are discrete values such as IP addresses, protocol names, and port numbers. To prepare it for machine learning analysis, it needs to undergo preprocessing steps. 
		
	\subsection{Data Preprocessing and Feature Engineering}\label{sec:preprocessing}
	Established preprocessing strategies for categorical data  aim to balance the trade-off between preserving important information and reducing the complexity of the data \cite{zheng2018feature}. Next we describe two methods suitable for our data and discuss their advantages and drawbacks. 
	 
	\subsubsection{Data Flattening }
	
	The flattened-data approach restructures network scan data by converting lists of ports and their associated services for each IP address into individual rows, creating a granular representation of IP-port-service combinations. Each row in the transformed dataset corresponds to a unique combination of an IP address, a port, and a service, with services converted into unique integer identifiers for efficient processing. This transformation represents in each row a single port and its corresponding service for a given IP.
	
	For example, a scan originally represented by one row with a list of ports and services is expanded into multiple rows, with each row representing a specific port-service pair for that scan. The flattening approach avoids the feature explosion associated with one-hot encoding while maintaining the relationships between IPs, ports, and services. The transformation can increase the number of rows of the dataset. For instance, transforming data for 100 IPs, 100 scans average per IP, and each scan with an average of 10 ports and 10 corresponding services, generates approximately 100,000 rows, while the raw data has only 10,000 rows. However, this 10-fold increase results from reformatting through unfolding of one list feature into 10 simple (integer) ones, not from computational inefficiency. 
	The main drawback of this method is that, while we need to identify anomalous IPs, the rows in this representation correspond to scans rather than IPs.
	
		 \begin{figure}
		\centering
		\includegraphics[width=0.6\textwidth]{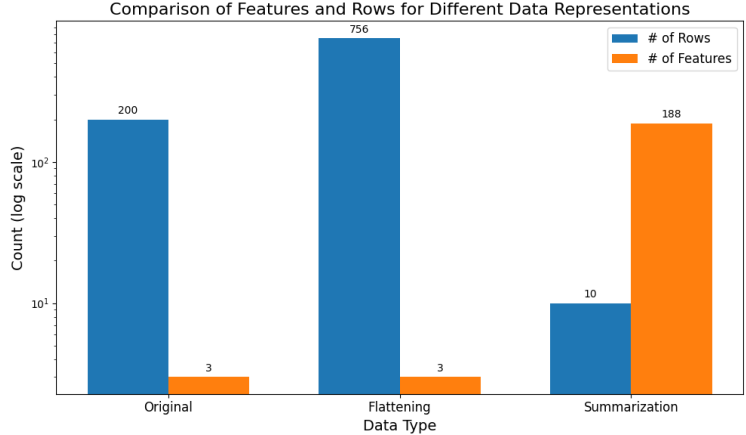}
		\caption{Comparison of the number of features and rows across three different data representations: original, flattening, and summarization. The original data contains 10 IP addresses with 20 scans each, resulting in 200 rows and 3 features (IP, port list, and service list). }
	\label{fig:datacounts}
\end{figure}
	
	\subsubsection{Summarization Approach}
	The summarization approach aggregates the data to create a fixed-length feature vector for every IP.	The method  restructures the data into a data frame with a row for each IP address, where each port and service is represented as a separate column, and the values provide the number of times the corresponding port or service are used for that IP. 
	While this approach reduces the number of rows (data points) required, it leads to a large number of columns. Since the number of possible services and ports of Censys data are currently 116 and 3552, respectively, the number of features could go up to more than 3600, even if a much smaller number of them are used per individual IP address. With such a high number of features, the risk of feature sparsity and overfitting increases. Another drawback of this approach is that it doesn't capture the correspondence between services and ports. While we do represent which ports and which services are used per IP, we don't encode how they match to each other. This can be avoided by having all possible  pairs (service, port) as features, but that will further increase the number of features by two orders of magnitude and will further exacerbate the problem.
	
	Figure~\ref{fig:datacounts} shows comparison between the number of datapoints and features of the above methods compared to those for the original dataset used in our experiments.

	\subsection{siForest Algorithm and Its Implementation}
	In network scan data, where features such as ports and services are represented as lists tied to IP addresses, retaining the structure of the data is essential. The proposed modification of the iForest algorithm addresses the  limitations of the original algorithm to work with such structured data. The core idea of the proposed method, siForest, is to halt further tree partitioning when all data points in a leaf node belong to the same IP address. The design ensures that related data points, such as the ports and services associated with an IP, remain grouped together, preserving their contextual relationships.
	
\begin{figure}
	\centering
	\includegraphics[width=0.9\textwidth]{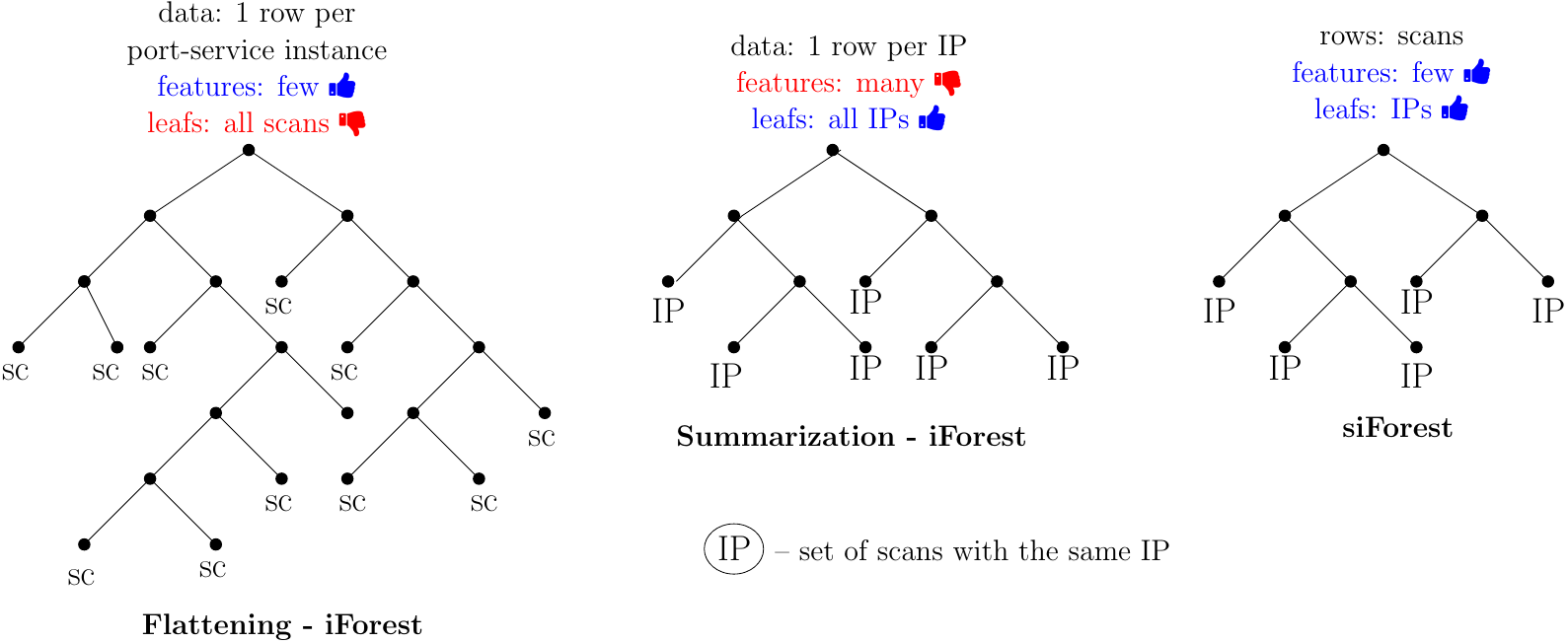}
	\caption{Comparison of Isolation Forest approaches showing datapoint types, features, and binary tree partitioning for the three methods considered in this paper.}
	\label{fig:trees-2}
\end{figure}

	This process is illustrated in Figure~\ref{fig:trees-2}, which shows three decision trees representing the different approaches considered in this paper. The first tree illustrates using iForest combined with the flattening method, which keeps the number of features low but does not allow the algorithm to identify anomalous IPs, as it operates at the scan level. The second tree, using the summarization method, aggregates scans into single IP-level data points, but this drastically increases the number of features and reduces the model’s efficacy. The third tree demonstrates the siForest approach, which uses scans as data points but stops splitting when a node contains only scans from the same IP. This maintains a low number of features while allowing it to detect anomalies at the IP level. It combines the advantages of the other two methods without their respective drawbacks.
	
	To adapt the iForest algorithm for set-structured data,  we made several  updates to the original scikit-learn implementation \cite{scikit-learn}. The most significant change is in the tree construction process, where siForest implements a stopping condition based on IP groupings. In the original iForest, partitioning continues until each leaf contains a single data point. However, siForest halts partitioning once all data points in a node belong to the same IP address, regardless of the number of the data points at the leaf.

	This adjustment  modifies the tree-building logic to evaluate whether all data points in a node share the same IP during splits. Additionally, the leaf structure is updated to store IP-specific data. This preserves the relationships between ports and services and the IP address they belong to, which are critical for detecting anomalies at IP level. 
	
	Another important update involves the scoring mechanism. Unlike the original implementation, where anomaly scores are calculated for individual data points, siForest aggregates scores at the IP level. We introduce two aggregation strategies: one that uses the minimum anomaly score among all scans for an IP and another using the average score. This allows siForest to capture both extreme outliers and persistent anomalous behavior. 
		
	\section{Experimental Analysis}
	In this section, we describe our procedure for comparing the performances of the different implementations of Isolation Forest described above and discuss the results.
	
		\subsection{Experiemntal Setup}
	
	We evaluated siForest against the traditional iForest algorithm using synthetic datasets that mirror real-world network traffic patterns. The datasets combine normal service-port pairings based on typical network scans with controlled anomalies to test detection capabilities under various configurations and preprocessing strategies. For data preprocessing, standard python libraries numpy \cite{numpy} and pandas \cite{pandas} are used.

	To create a realistic testing environment, the synthetic data incorporates both normal behavior and injected anomalies. Normal patterns reflect common service-port combinations observed in network scans, while anomalies include usage spikes, non-standard port assignments, and service-port mismatches. This controlled environment enables consistent testing across scenarios.
	
	We compare the three methods discussed in \cref{sec:preprocessing} to address the set-structured nature of network scan data. We use the implementation of iForest on scikit-learn package \cite{scikit-learn} and our implementation of siForest as described above. Both algorithms use the same number of 100 estimators (trees) for consistency. We calculate anomaly scores at the IP level using the minimum score option as it gives a better performance for the generated data.
		
	Data is generated for each of the two anomaly types, with 10 different random sets for each type. The three tested algorithms are then run on the same data, and their predictions are compared with the known (planted) anomalies. Performance is evaluated using standard metrics such as precision, recall, and F2-score. Although precision and F2-score provide valuable information that should be taken into account, recall is most relevant because attack surface identification primarily aims to avoid missing potential vulnerabilities. Detected anomalies are inspected by a human expert to determine if they may pose a real threat. However, if a vulnerability is missed by the anomaly detector, it will never be analyzed or addressed.		
		
	\subsection{Data Generation}
	While trying to mimic Censys \cite{censys} collected network data, our study uses synthetic dataset to enable controlled testing of anomaly detection methods. We base normal network behavior on common service-port combinations documented in Censys, such as HTTP on port 80 and FTP on port 21. A Python dictionary stores these standard mappings to maintain consistency in generating baseline network activity.

	A synthetic data approach allows precise injection of anomalies through deviations from expected service-port relationships. 
	The evaluation covers two distinct anomaly classes, described below, to test how each method's detection capabilities vary depending on the anomaly type. 
	
	This design choice overcomes the limitation of unlabeled anomalies in raw Censys data while allowing flexibility in defining both "normal" and "anomalous" activity \cite{barnett1974outliers}.
	
	\subsection{Anomaly Types}

\begin{figure}
	\centering
	\includegraphics[width=0.7\linewidth]{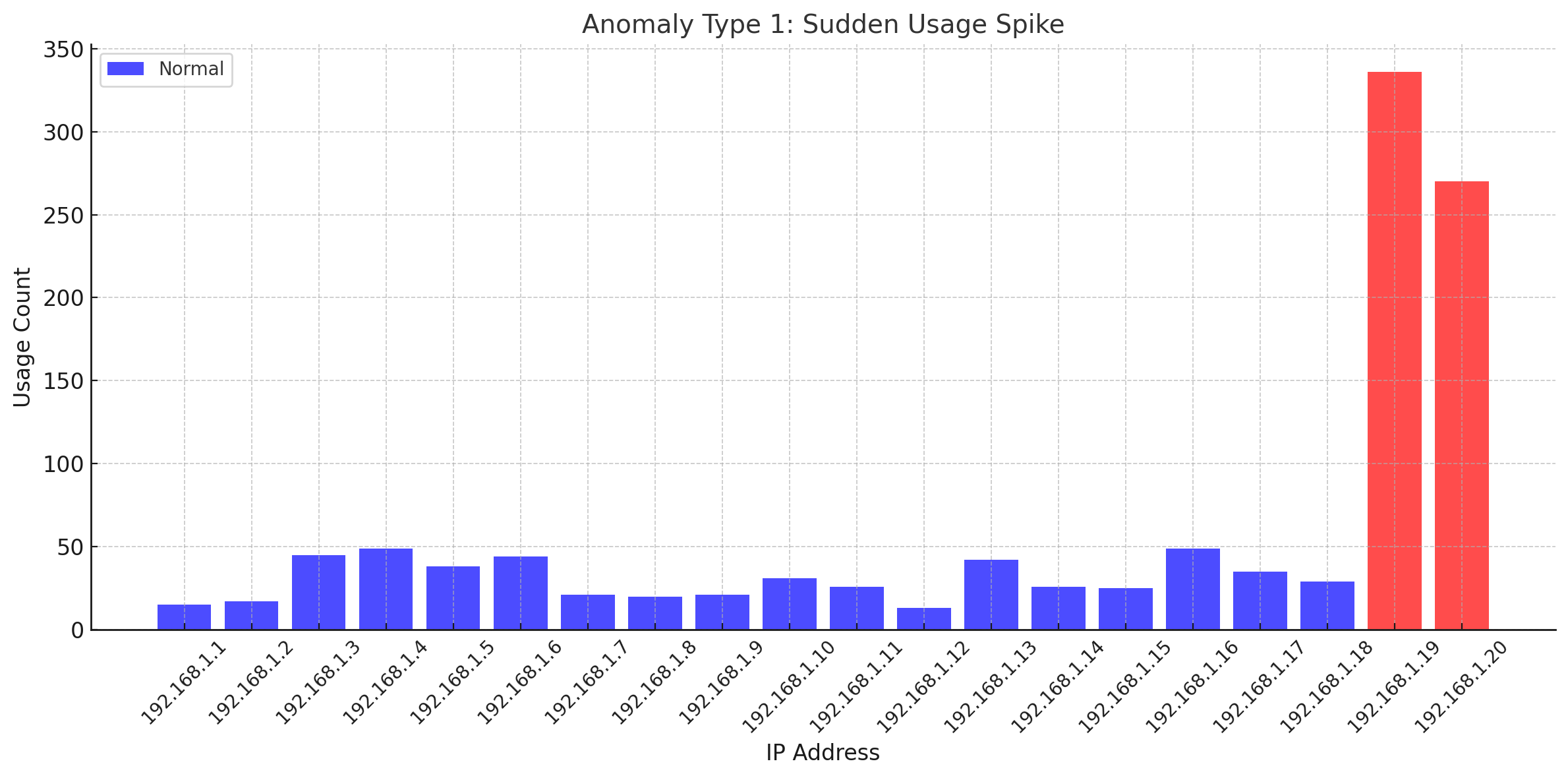}
	\caption{Bars in red indicate anomalous IP scans with much higher usage compared to normal scans.}
	\label{fig:usage}
	\end{figure}

	Our evaluation incorporates two distinct anomaly patterns in the synthetic dataset. The first type, referred to as anomaly type 1,  represents sudden usage spikes, where an IP address shows unexpected activity across one or more ports \cite{fortinet_anomaly}. Such patterns often indicate network scanning or denial-of-service attempts. Figure \ref{fig:usage} shows these volume-based anomalies, with red bars marking abnormal scanning activity against the blue baseline.

\begin{figure}
	\centering
	\includegraphics[width=0.7\linewidth]{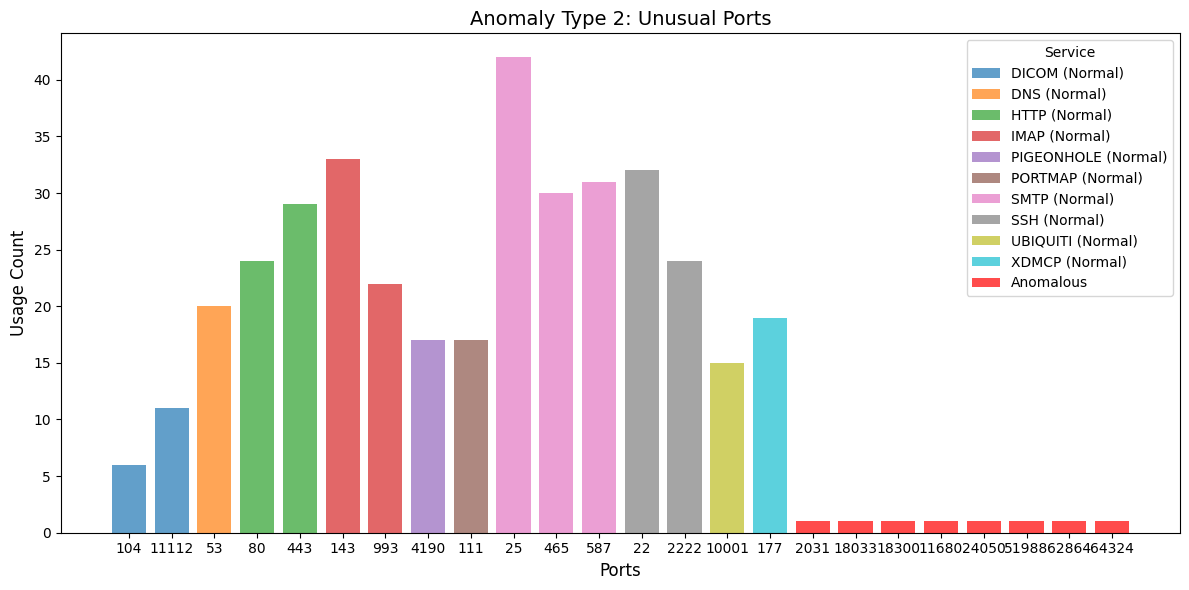}
	\caption{Red bars highlight anomalous behavior based on atypical port usage.}
	\label{fig:unusual}
	\end{figure}

	The second type, referred to as anomaly type 2,  introduces non-standard service-port combinations \cite{howtogeek_ports}, such as HTTP services running on unexpected ports. Figure \ref{fig:unusual} illustrates these configuration anomalies, with red segments showing services on non-standard ports. By generating these irregular pairings, the dataset evaluates the model's capability to detect anomalies that arise from misconfigurations or intentional misuse of network resources. 

	The frequency and intensity of both anomaly types can be adjusted using tunable parameters to create diverse testing scenarios, from subtle deviations to clear outliers, enabling a thorough assessment of each model's detection sensitivity across different anomaly distributions. 

	\subsection{Performance Comparison}
\begin{figure}
		\centering
		\includegraphics[width=0.95\textwidth]{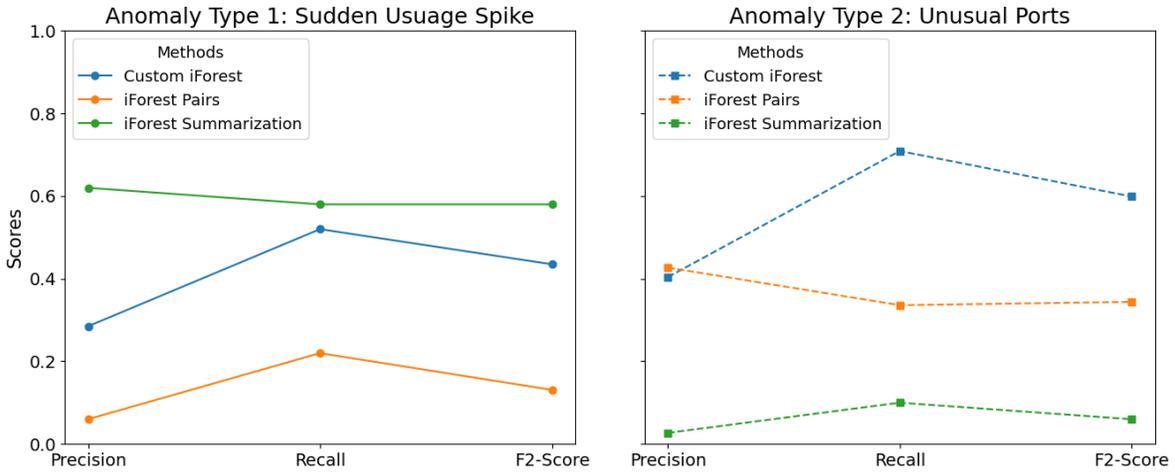}
		\caption{Comparison of detection performance across anomaly types and methods.}
		\label{fig:scores}
	\end{figure}

	The evaluation results, shown on Figure \ref{fig:scores}, reveal distinct performance patterns across anomaly types and detection methods. For type 1 anomalies (usage spikes),  siForest demonstrated robust performance with a precision of 0.285, recall of 0.520, and F2-score of 0.435. This balanced detection capability suggests the method effectively identifies volume-based anomalies while maintaining reasonable false positive rates.
	
	The flattening preprocessing approach, implemented through iForest Pairs, shows significant limitations with type 1 anomalies. Its precision (0.060), recall (0.220), and F2-score (0.131) are substantially lower than the other methods, likely due to the loss of structural information during the flattening process. In contrast, iForest with summarization preprocessing achieved the highest precision (0.61) among all approaches while maintaining strong recall (0.58), resulting in an F2-score of 0.59. This performance indicates that summarization effectively preserves critical patterns for type 1 anomalies. It seems the summarized data approach successfully captures unusual behavior for type 1 anomalies because it aggregates and counts occurrences, aligning with the volume-based nature of this anomaly type characterized by increased usage patterns.
	
	For type 2 anomalies (unusual port usage), iForest Summarization  struggles, achieving only 0.027 precision, 0.100 recall, and an F2-score of 0.060. This sharp decline in performance suggests that the summarization strategy is particularly ineffective for detecting this type of anomaly. In comparison, Custom iForest demonstrates the strongest performance for type 2 anomalies, with a precision of 0.404, recall of 0.709, and F2-score of 0.599. These results, particularly the high recall, suggest the method excels at identifying structural anomalies in service-port relationships.
	
	The flattening approach (iForest pairs) shows better performance for type 2 anomalies compared to type 1, with a precision of 0.427, recall of 0.336, and F2-score of 0.344. While still underperforming compared to Custom iForest, this approach demonstrates less dramatic degradation compared to its performance on type 1 anomalies.
	
	The two plots reveal distinct performance characteristics across different preprocessing strategies. Summarization preprocessing dominates in precision for type 1 anomalies but performs poorly for type 2 anomalies. Custom iForest (siForest) achieves consistently good performance across both anomaly types -- it shows the best overall performance for type 2 anomalies and maintains strong results for type 1 anomalies, particularly in recall. 
	
	\section{Conclusions and Future Work}
	This paper discusses the usage of the Isolation forest approach to detecting anomalies in internet scan data for applications in attack surface identification. We are applying two data preprocessing methods  and also develop a modification of the standard iForest algorithm that is able to work with a set-structured data. Our analysis shows that the effectiveness of the methods depend on the type of anomaly being present. While the summarization method particularly excels at identifying volume-based anomalies, it performs poorly when detecting unusual port usage patterns. The flattening approach, on the other hand, performs poorly on volume-based anomalies, but beats the summarization one for anomalies based on unusual port usage.

	The proposed siForest algorithm demonstrates the most consistent performance across both anomaly types. For unusual port usage, it achieves the best overall performance, while for the other type of anomaly its performance comes close to that of the summarization approach, especially with respect to the important recall metrics.

	These results suggest that proper preprocessing methods can help detect anomalies in set-structured datasets such as ones based on Censys internet scans, but siForest is the only methods among the tested ones that shows consistently good performance over all anomaly types.

	This work leaves several open avenues for future research. One is to test the algorithms on more anomaly types and on datasets of varying sizes. Applying siForest to real-world datasets is also very interesting although such data may be hard to obtain. Integrating graph-based techniques to capture more intricate relationships within network data is another promising direction, as is the development of automated mechanisms for anomaly characterization \cite{chalapathy2019deep}. 
	
	By broadening its scope and improving its scalability, SiForest has the potential to develop into a practical tool for anomaly detection in set-structured data, and in cybersecurity and attack surface identification, in particular.

\bibliographystyle{abbrv}

\end{document}